\documentclass[sigconf]{acmart}

\AtBeginDocument{%
	\providecommand\BibTeX{{%
			\normalfont B\kern-0.5em{\scshape i\kern-0.25em b}\kern-0.8em\TeX}}}

\copyrightyear{2021}
\acmYear{2021}
\setcopyright{acmcopyright}\acmConference[SIGSPATIAL '21]{29th International Conference on Advances in Geographic Information Systems}{November 2--5, 2021}{Beijing, China}
\acmBooktitle{29th International Conference on Advances in Geographic Information Systems (SIGSPATIAL '21), November 2--5, 2021, Beijing, China}
\acmPrice{15.00}
\acmDOI{10.1145/3474717.3488239}
\acmISBN{978-1-4503-8664-7/21/11}

\begin{document}
	
	\title{Multi-View Spatial-Temporal Model for Travel Time Estimation}
	\author{Zichuan Liu}
	\affiliation{%
		\institution{Wuhan University of Technology}
		\city{Wuhan}
		\country{China}
	}
	\email{lzc775269512@gmail.com}
	
	\author{Zhaoyang Wu}
	\affiliation{%
		\institution{East China Normal University}
		\city{Shanghai}
		\country{China}
	}
	\email{51195100017@stu.ecnu.edu.cn}
	
	\author{Meng Wang}
	\affiliation{%
		\institution{Sun Yat-sen University}
		\city{Guangzhou}
		\country{China}
	}
	\email{wangm329@mail2.sysu.edu.cn}

	\author{Rui Zhang*}
	\affiliation{%
		\institution{ Wuhan University of Technology}
		\city{Wuhan}
		\country{China}
	}
	\email{zhangrui@whut.edu.cn}

	\renewcommand{\shortauthors}{Liu, et al.}
	
	\begin{abstract}
		Taxi arrival time prediction is essential for building intelligent transportation systems. Traditional prediction methods mainly rely on extracting features from traffic maps, which cannot model complex situations and nonlinear spatial and temporal relationships. Therefore, we propose Multi-View Spatial-Temporal Model (MVSTM) to capture the mutual dependence of spatial-temporal relations and trajectory features. Specifically, we use graph2vec to model the spatial view, dual-channel temporal module to model the trajectory view, and structural embedding to model traffic semantics. Experiments on large-scale taxi trajectory data have shown that our approach is more effective than the existing novel methods. The source code can be found at https://github.com/775269512/SIGSPATIAL-2021-GISCUP-4th-Solution.
	\end{abstract}

	\ccsdesc[500]{Computing methodologies~Modeling methodologies}
	\keywords{multi-view, travel time estimation, spatial-temporal, graph2vec}
	
	
	\maketitle
	
	\section{Introduction}

	Travel Time Estimation (TTE), also known as Estimated Time of Arrival (ETA), is the most important, complex, and challenging problem in intelligent transportation systems and location-based information services. The estimated time of arrival can help the platform make better decisions in such scenarios as traffic monitoring \cite{chawla2012inferring}, carpooling \cite{ma2013t} and taxi dispatch \cite{yuan2012t}. However, arrival time can be affected by route length, real-time road conditions, traffic lights and other factors. Therefore, how to extract an effective road network structure characterization is indispensable for travel time prediction.
	
	Currently, some researches \cite{asghari2015probabilistic, li2019learning} focus on path-based methods in which matrix factorization technology is used to estimate travel time. It can estimate the travel time of all roads under different conditions in a period. In \cite{prokhorchuk2019estimating}, the authors use Bayesian networks to estimate distributions of travel times which can infer travel time distribution from sparse GPS measurement data. In \cite{meng2017link}, Meng et al. introduce a likelihood function and estimate the most likely traffic delay for each road segment by maximizing the function. Although travel time segmentation is computationally simple and intuitive, yet the number of errors will increase as the travel length extends. This is because that they fail to match the road network to a specific trajectory or consider the impact of other road sections on one road section. Wang et al. \cite{wang2014travel} combine the geospatial, temporal and historical background with mapping data learned from the trajectory. However, they ignore the dynamic changes in a real-time traffic network. In \cite{wang2018will}, Wang et al. propose spatial correlation operations by integrating geographic information into classical convolution and design an end-to-end deep learning framework for travel time estimation (DeepTTE). However, it does not capture the upstream and downstream nodes of the trajectory change caused by vehicle movements or exploit the information on the trajectory maps. Fu et al. \cite{fu2019deepist} use convolution neural network to process map image trajectory information, which combines traffic mode recommendation with travel time estimation recommendation. Similarly, their methods are not multi-view, and there is a lack of spatial-temporal dependence.

	In this paper, we propose a multi-view temporal model (MVSTM), which jointly considers the relationship between space, time, and nodes. First, we embed the learning area and the latent semantics of the trajectory through the graph. Then, we use LSTM \cite{hochreiter1997long} and attention \cite{vaswani2017attention} dual-channel prediction to capture dynamic historical trajectory features. Finally, the static road features of the road section are extracted by the structural embedding method. We have conducted extensive experiments on a large-scale taxi trajectory dataset from Didi Shenzhen. The results show that our method consistently outperforms the competing baselines.
	
	\section{Preliminaries}
	
	In this section, we first define some symbols and formulate the problem of travel time estimation. Briefly, we use the index of time intervals $t$ to represent real-time traffic conditions $C_t$ whose factors are similar to the road network and weather. A travel trajectory $T_i$ is composed of link parts $l_n$ and crossing parts $c_m$. These parts are organized in sequences, i.e. $T_1 = \left\lbrace l_1, l_2, . . ., l_n, c_1, c_2, . . ., c_m\right\rbrace _1$, and every element of the sequence is composed by the data fields. Each link has $j$ categorical features, (e.g. \textit{link\_id} and \textit{link\_current\_ status}) and $k$ numerical features, (e.g. \textit{link\_time} and \textit{link\_ratio}).
	Each categorical feature is mapped to vector $l^{cat} \in \mathbb{R}^{d\times1}$ by using the embedding method, and each numerical feature is transformed to vector $l^{num} \in \mathbb{R}^{d\times1}$ by applying a linear transformation.
	Therefore a link $l_i$ is defined as $l_i = l^{cat_0} \oplus \cdots \oplus l^{cat_{j-1}} \oplus l^{num_0} \oplus \cdots \oplus l^{num_{k-1}}$, where $\oplus$ denotes element-wise addition. Similarly, crossing parts $c_i$ are also defined as above. Moreover, each piece of GPS data contains some global information, such as distance and departure time slice as the head information $h_i$ of the trajectory $T_i$. The task here is to predict the arrival time of a travel trajectory. Therefore, our final goal is to predict:
	\begin{gather}
		eta_i = \mathbb{F} ( T_i, h_i, C_t )
	\end{gather}
	where $eta_i$ is the estimated time of arrival of the testing $T_i$ and $ata_i$ is the real travel time. We define our prediction function $\mathbb{F}(\cdot )$ to capture the complex spatial and temporal interaction between travel trajectories and traffic conditions.

	\section{MODEL ARCHITECTURE}
\begin{figure*}
		\includegraphics[width=.8\textwidth]{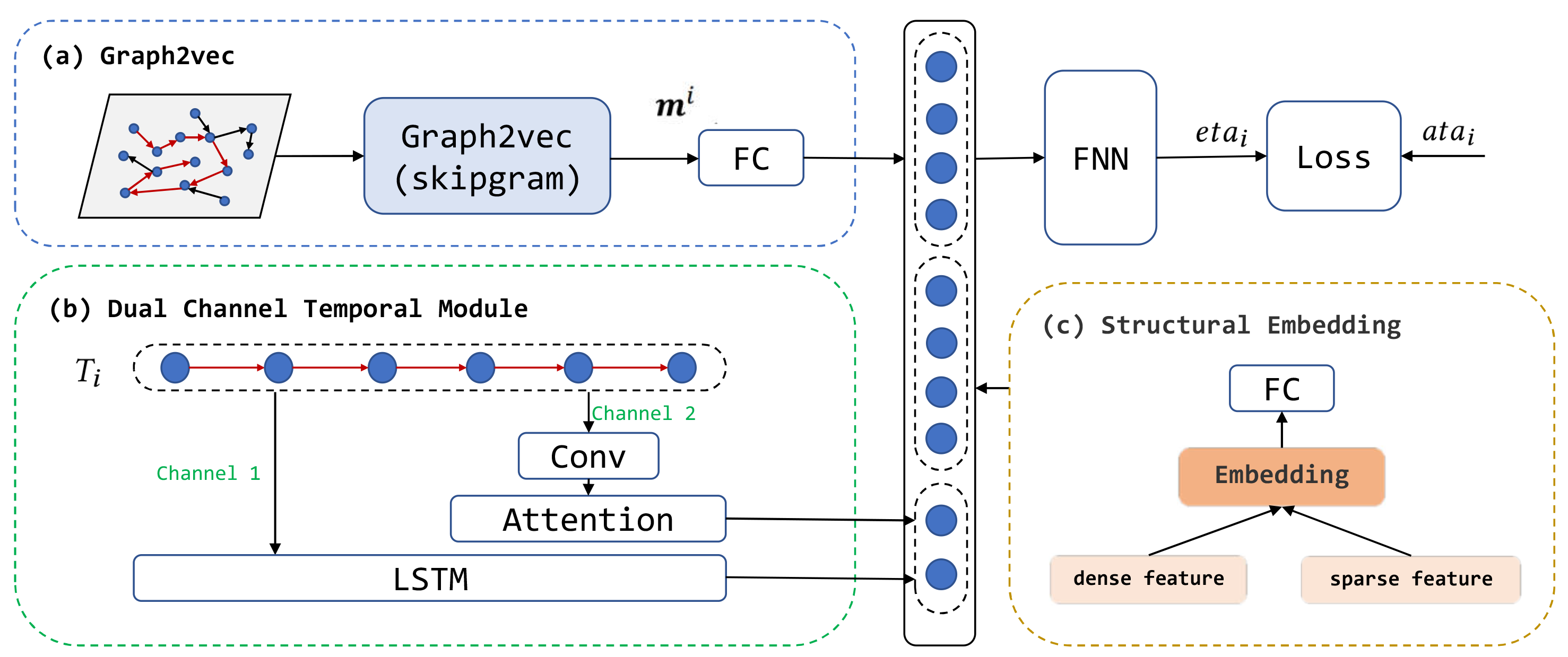}
		\caption{The Architecture of MVSTM. (a) The spatial component uses graph2vec to capture the spatial dependence between the regions near the trajectory and represent the geographic information of the trajectory. (b) The trajectory view employs a dual-channel structure. For each trajectory, the first channel uses LSTM to capture the historical information of the vehicle passing through the node. The second channel first uses CNN to generalize the features, and then uses the attention mechanism to capture the long-term time dependence. (c) The semantic view embeds the global features of the trajectory on the whole map into vectors to describe the global trajectory. Finally, the fully connected layers are used for joint training and prediction.} 
		\label{fig1}
	\end{figure*}

	In this section, we describe the structure of Multi View Spatial-Temporal Model (MVSTM). Figure \ref{fig1} shows the architecture of our proposed method, including three views: spatial view, trajectory view and semantic view. 
	
	\subsection{Spatial View: Graph2vec}
	For the road representation of a trajectory on the entire map, the spatial information of the trajectory during this period of time can be mined. Recent works \cite{hong2020heteta, fang2020constgat} on representation learning for graph-structured maps mainly focus on embedding representations of entire graph structures via all edges and nodes. However, for some networks that contain millions of nodes, it will take massive amounts of time and memory to represent the whole graph as fixed-length feature vectors. To address this limitation, in this section, we follow the graph2vec of Narayanan et al. \cite{narayanan2017graph2vec} to represent the trajectory’s spatial information in the subgraph.

	As shown in Figure \ref{fig1} (a), at each time interval $t$, we represent the location information $G_t$ of the upstream and downstream road network of a certain trajectory in the whole map. In graph2vec, we replace the road graph analogy with a document composed of root subgraphs, which are analogous words from a special language. More specifically, given a set of subgraphs $S = \left\lbrace G_1, G_2, . . ., G_n\right\rbrace$
	and a sequence of nodes $n(G_t) = \left\lbrace w_1, w_2, . . ., w_l\right\rbrace _t$ sampled from subgraphs $G_t \in S$, graph2vec
	skipgram learns a $\delta $ dimensional embeddings of the set of subgraphs $G_t \in S$ and each nodes $w_j$ sampled from $n(G_t)$. The module works by
	considering a node $w_j \in n(G_t)$ to be occurring in the context of subgraph $G_t$ and tries to maximize the following log-likelihood:
	\begin{gather}
		\sum_{j=1}^{l}log Pr(w_j|G_t)
	\end{gather}
	where, the probability $Pr(w_j|G_t)$ is defined as,
	\begin{gather}
		\frac{exp(\vec{G}\cdot \vec{w_j})}{\sum_{w\in V} exp(\vec{G}\cdot \vec{w})}
	\end{gather}
	where $V$ is the vocabulary of all the nodes in the traffic map. Finally, the spatial feature $m_i$ generated by graph coding is used to represent the context information of the trajectory.

	\subsection{Trajectory View: Dual Channel Temporal Module}
	
	As links are traversed by cars sequentially, we choose Long-Short Term Memory (LSTM) \cite{hochreiter1997long} to extract temporal information. LSTM, as a type of recurrent neural network (RNN), has been successfully used in many fields, such as NLP \cite{bahdanau2014neural} and time series prediction \cite{siami2018comparison}. In a period, a trajectory sequence $T_i$ is used as input and updated as:
	
	\begin{equation}
	\begin{aligned}
	\mathbf{i}_{t} &=\sigma(\mathbf{W}_{i}\left[\mathbf{x}_{t} ; \mathbf{h}_{t-1}\right]+\mathbf{b}_{i}), \\
	\mathbf{f}_{t} &=\sigma(\mathbf{W}_{f}\left[\mathbf{x}_{t} ; \mathbf{h}_{t-1}\right]+\mathbf{b}_{f}), \\
	\mathbf{o}_{t} &=\sigma(\mathbf{W}_{o}\left[\mathbf{x}_{t} ; \mathbf{h}_{t-1}\right]+\mathbf{b}_{o}), \\
	\mathbf{g}_{t} &=\tanh (\mathbf{W}_{g}\left[\mathbf{x}_{t} ; \mathbf{h}_{t-1}\right]+\mathbf{b}_{g}),
	\end{aligned}
	\end{equation}
	where $\sigma$ is a sigmoid activation function and $\mathbf{W}_{i}$, $\mathbf{W}_{f}$, $\mathbf{W}_{o}$, $\mathbf{W}_{g}$ are affine transformations. By feeding $T_i \in \mathbb{R}^{n \times d}$ to LSTM and selecting the last hidden state, we get a feature vector $h_l$ representing the whole trip.
	
	However, a potential issue with this approach is that it has to compress all the information of links into a fixed-length vector \cite{bahdanau2014neural}. To fix this problem, we create a second channel feature using an attention mechanism. The attention mechanism is a vital part of Transformer \cite{vaswani2017attention} which is used as the building block by many state-of-the-art models. Unlike LSTM, the attention mechanism has no inductive bias in the order of road links. Moreover, it can process input in parallel, and mine links relation when they are far apart. Attention can be expressed as:
	
	\begin{equation}
	\operatorname{Attention}(Q, K, V)=\operatorname{softmax}\left(\frac{Q K^{T}}{\sqrt{d_{k}}}\right) V
	\end{equation}

	As demonstrated in \cite{rakhlin2016convolutional}, a simple convolutional neural network can improve the state-of-the-art on many NLP tasks. So we first apply a 1D convolution $w \in \mathbb{R}^{h*d}$ to $T_i$. We set the number of filters to $d$ and add paddings before convolution to get a \textit{feature map} $c$ with the same shape as $T_i$. After applying 1D convolution, we put $c$ through the self-attention layer where \textit{query} \textit{key}, and \textit{value} equals $c$,
	then sum up the output over the link length axis, resulting in a vector $h_a \in \mathbb{R}^{h \times 1}$. Finally, we combine the results from the two channels and represent them as the features of the trajectory.

	\subsection{Semantic View: Structural Embedding}
	
	Apart from link-related features, other features have an influence on ETA, including the distance of the path, simple eta, time slice, driver id, day of the week, weather and temperature in $h_i$ and $C_t$. For each categorical feature, we transform it to a vector through the corresponding embedding layer and then concatenate these vectors and numerical features to a combined vector $S_i$ as:
	\begin{gather}
		S_i = h_i^{cat_1} \oplus \cdots \oplus h_i^{cat_{j}} \oplus C_t^{cat_1} \oplus \cdots \oplus C_t^{cat_k}
	\end{gather}
	where $\oplus$ denotes element-wise addition. 
	
	\subsection{Loss function} MVSTM is trained by minimizing the loss
	function of mean absolute percentage error (MAPE) for jointly training our proposed model. We define the loss function between the predicted results $eta_i$ and the real data $ata_i$ as:
	\begin{gather}
		L(\theta ) = \frac{1}{n}\sum_{i=0}^{n}\frac{\left | eta_i-ata_i \right |}{ata_i }
	\end{gather}
	where $\theta$ denotes the learnable parameters in our proposed model.

	\section{Experiment}
	
	\subsection{Datasets and Metric}
	
\begin{figure}
		\includegraphics[width=.5\textwidth]{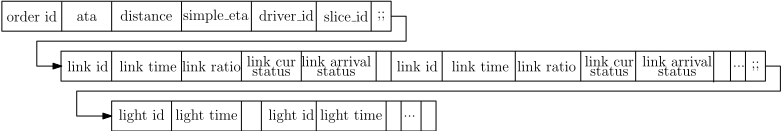}
		\caption{Sketch map is composed of track head, link part and crossing part of each track.} 
		\label{fig2}
	\end{figure}

	\textbf{Dataset: } In this paper, we use a large-scale taxi request data set collected in Shenzhen City by Didi Chuxing, one of the largest online taxi-hailing companies in China. The training dataset contains taxi trip information from August 1st to August 31st and the test dataset contains trip information of September 1st. The schematic diagram of each track is shown in \ref{fig2} \footnote{https://sigspatial2021.sigspatial.org/images/sketchmap.svg}, which contains a road network (map) and weather data of the day.
	
	\textbf{Metric: } We use average percentage error (MAPE) to evaluate our algorithm, which is defined as follows:
	\begin{gather}
		MAPE = \frac{1}{n}\sum_{i=0}^{n}\frac{\left | \hat{y_i}-y_i\right |}{y_i}
	\end{gather}
	
	where $y_i$ and $\hat{y_i}$ mean the real value and predicted value of arrival time of the taxi $i$, respectively. $n$ denotes the total number of trajectories.
	
	\subsection{Methods for Comparison}
	We compared our model with the following methods by adjusting the parameters for all methods.

	\begin{itemize} 
		\item \textbf{ Simple ETA:} Simple ETA is the accumulation of average link time of departure link time and cross time.
		
		\item \textbf{LightGBM \cite{ke2017lightgbm}:} LightGBM is a powerful tree-based boosting method, which is widely used in data mining applications.
		
		\item \textbf{Multiple layer perceptron (MLP):} We set the number of hidden units of the three-layer fully connected neural network to 512, 128, and 64, respectively.

		\item \textbf{DeepFM \cite{guo2017deepfm}:} DeepFM is the most classic model in recommendation systems. We apply it to our semantic view.
		
		\item \textbf{xDeepFM \cite{lian2018xdeepfm}:} XDeepFM is an upgraded version of DeepFM which proposes CIN to embed sparse features. It is also commonly used in some scenarios of recommendation systems.
		
		\item \textbf{WDR \cite{wang2018learning}:} The core idea of WDR model is a global model and recurrent model. The function of the global model is similar to DeepFM, which learns the global information of travel. The recurrent model focuses on learning local details such as link sequences.
		
	\end{itemize}
	
	We also analyzed the impact of different view components proposed in our model.

	\begin{itemize} 
		\item \textbf{ Trajectory + Semantic view:} For this variant, we only use trajectory input as a dual-channel module and connect the global information after embedded as output.

		\item \textbf{Spatial + Trajectory(RNN) view:} This variant
		considers both spatial and trajectory views. However, we replaced trajectory views and used a single RNN channel LSTM \cite{hochreiter1997long} as the recurrent model.

		\item \textbf{MVSTM:} Our proposed method, which combines spatial, trajectory and semantic views.
		
	\end{itemize}
	
	\subsection{Performance Comparison}
	Table \ref{tab1} shows that the proposed method has the best performance compared with all the other methods. More specifically, we can see that LightGBM and MLP are not embedding sparse features, and their performance is poor. The method in the recommender system takes sparse features into account and therefore achieves better results. At present, the state-of-the-art method WDR uses global features and trajectory sequences as research, and achieves better results than traditional regression models. However, the above-mentioned methods do not mine the spatial information of trajectory in the traffic network. With the addition of semantic view (Graph2vec), most methods show performance improvements. Therefore, our proposed method outperforms these methods.
	
	Furthermore, our proposed method achieves 12.2024\% (MAPE). It can be seen that different views improve the results. Our final submission scheme integrates MVSTM and LightGBM, with weights of 0.9 and 0.1 respectively, and the final result is 0.12177.

	\begin{table}
		\caption{Comparison with Different Baselines}
		\label{tab1}
		\begin{tabular}{c|c}
			\toprule
			Method and Component Analysis &MAPE\\
			\midrule
			Simple ETA	 & 0.16368\\
			LightGBM & 0.15742\\
			Graph2vec + LightGBM & 0.15485 \\
			MLP & 0.15582\\
			Graph2vec + MLP & 0.14834\\
			DeepFM & 0.14945\\
			xDeepFM & 0.14921\\
			Graph2vec + xDeepFM & 0.14627\\
			WDR & 0.12831\\
			\bottomrule
			\textbf{ Trajectory + Semantic view} (ours) &\textbf{ 0.12654}\\
			\textbf{Spatial + Trajectory(RNN) view} (ours) & \textbf{0.12547}\\
			\textbf{MVSTM} (ours) & \textbf{0.12202}\\
			\bottomrule
		\end{tabular}
	\end{table}

	\section{Conclusion and Discussion}
	In this paper, we propose a travel time estimation model to improve the accuracy of travel time estimation. In order to overcome the shortcomings of existing work, we design Multi-View Spatial-Temporal Model (MVSTM) to represent spatial information, trajectory information and semantic information. In future research, we will consider using some novel spatial-temporal map methods and adding POI interest nodes to represent travel.

	\bibliographystyle{ACM-Reference-Format}
	\bibliography{sample-base}

\end{document}